\documentclass[10pt,twocolumn,letterpaper]{article}

\usepackage{cvpr}              

\usepackage{graphicx}
\usepackage{amsmath}
\usepackage{amssymb}
\usepackage{booktabs}

\usepackage{graphicx}
\usepackage{array}
\usepackage{booktabs}
\usepackage{url}
\usepackage{multirow}
\usepackage{placeins}
\usepackage{amsmath}
\usepackage{enumitem}

\newcolumntype{L}{>{\arraybackslash}m{4cm}}
\newcolumntype{V}{>{\arraybackslash}m{6.5cm}}
\newcolumntype{M}{>{\centering\arraybackslash}m{2.5cm}}
\newcolumntype{A}{>{\centering\arraybackslash}m{4cm}}
\newcolumntype{N}{>{\arraybackslash}m{3cm}}
 
\newcolumntype{S}{>{\centering\arraybackslash}m{1cm}}
\newcolumntype{X}{>{\arraybackslash}m{1cm}}
\newcolumntype{P}{>{\arraybackslash}m{10cm}}
\newcolumntype{Q}{>{\arraybackslash}m{5cm}}
\newcolumntype{B}{>{\centering\arraybackslash}m{4cm}}
\newcolumntype{K}{>{\centering\arraybackslash}m{5cm}}
\usepackage{color, colortbl}

\definecolor{Orange}{rgb}{1.0, 0.27, 0.0}
\definecolor{Green}{rgb}{0.0, 1.0, 0.0}
\definecolor{Blue}{rgb}{0.0, 0.0, 1.0}
\NewDocumentCommand{\lingji}{ mO{}}{\textcolor{Orange}{\textsuperscript{\textit{lingji}}\textsf{\textbf{\small[#1]}}}}
\NewDocumentCommand{\chengjie}{ mO{}}{\textcolor{Green}{\textsuperscript{\textit{chengjie}}\textsf{\textbf{\small[#1]}}}}
\NewDocumentCommand{\cezhang}{ mO{}}{\textcolor{Blue}{\textsuperscript{\textit{ce}}\textsf{\textbf{\small[#1]}}}}

\usepackage[accsupp]{axessibility} 

\usepackage[pagebackref,breaklinks,colorlinks]{hyperref}

\usepackage[capitalize]{cleveref}
\crefname{section}{Sec.}{Secs.}
\Crefname{section}{Section}{Sections}
\Crefname{table}{Table}{Tables}
\crefname{table}{Tab.}{Tabs.}

\def\cvprPaperID{*****} 
\def\confName{CVPR}
\def\confYear{2023}

\begin{document}

\title{MotionTrack: End-to-End Transformer-based Multi-Object \\Tracking with LiDAR-Camera Fusion}

\author{Ce Zhang\thanks{Work conducted during an internship at Motional}\\ 
Virginia Tech\\
{\tt\small zce@vt.edu}
\and
Chengjie Zhang\\
Motional\\
{\tt\small chengjie.zhang@motional.com}
\and
Yiluan Guo\\
Motional\\
{\tt\small yiluan.guo@motional.com}
\and
Lingji Chen\\
Motional\\
{\tt\small lingji.chen@motional.com}
\and
Michael Happold\\
Motional\\
{\tt\small michael.happold@motional.com}
}
\maketitle

\begin{abstract}

Multiple Object Tracking (MOT) is crucial to autonomous vehicle perception. 
End-to-end transformer-based algorithms, which detect and track objects simultaneously, show great potential for the MOT task. 
However, most existing methods focus on image-based tracking with a single object category. 
In this paper, we propose an end-to-end transformer-based MOT algorithm (MotionTrack) with multi-modality sensor inputs to track objects with multiple classes.
Our objective is to establish a transformer baseline for the MOT in an autonomous driving environment. 
The proposed algorithm consists of a transformer-based data association (DA) module and a transformer-based query enhancement module to achieve MOT and Multiple Object Detection (MOD) simultaneously. 
The MotionTrack and its variations achieve better results (AMOTA score at 0.55) on the nuScenes dataset compared with other classical baseline models, such as the AB3DMOT, the CenterTrack, and the probabilistic 3D Kalman filter. 
In addition, we prove that a modified attention mechanism can be utilized for DA to accomplish the MOT, and aggregate history features to enhance the MOD performance.

\end{abstract}

\section{INTRODUCTION}
Perception is a fundamental and key element for autonomous vehicles. 
Common perception tasks fall into three categories  \cite{8936542}: Multiple Object Detection (MOD), Multiple Object Tracking (MOT), and Multiple Object Prediction (MOP).
A reliable MOT algorithm shall comprehend the MOD outcomes and establish a connection for the MOP. 

\begin{figure}[ht!]
\centering
\includegraphics[width=8.2cm]{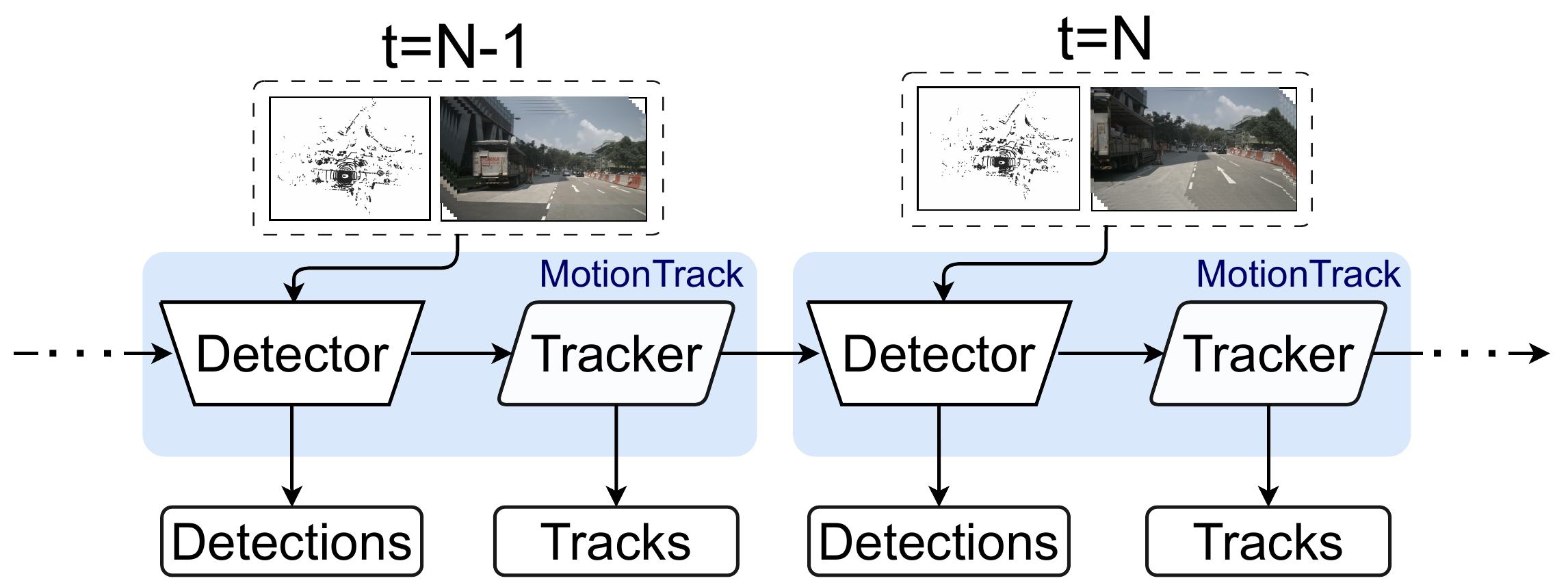}
\caption{MotionTrack Model Demonstration} 
\label{fig:demo1}
\end{figure}

Machine learning-based (ML-based) tracking algorithms recently become popular to improve MOT performance by enhancing the temporal and spatial features through learning \cite{nam2016learning,9093977,lukezic2017discriminative}.
Current ML-based tracking algorithms have two paradigms: tracking by detection, and simultaneous tracking and detection \cite{Bashar2022MultipleOT}.
The former considers detection and tracking as separate and sequential tasks, while the latter jointly processes detection and tracking at the same time.
Both paradigms utilize neural networks for motion prediction (MP) or data association (DA).
Simultaneous tracking and detection offer a significant advantage through mutual feature sharing.
Specifically, temporal and spatial features from tracking can improve detection performance, whereas appearance and position features from detection can enhance DA in tracking.
Because of these benefits, we choose the simultaneous tracking and detection paradigm for MotionTrack.

The transformer architecture and the attention mechanism, originally applied in the field of natural language processing \cite{vaswani2017attention}, perform well for vision tasks such as object detection, image quality assessment, and pose estimation \cite{Singh2023TransformerBasedSF, zhang2022quality, zhang2021attention, 9812060, zhang2022monodetr, xu2022vitpose, 10.1007/978-3-031-25066-8_41, Singh2023SurroundViewV3}. 
Recent studies indicate that the transformer model can be utilized for tracking tasks by estimating object motion and transferring appearance and motion features \cite{10.1007/978-3-031-19812-0_38, 9880137, 9879668, sun2020transtrack, Ruppel2022TransformersFM}. 
In view of the nature of the transformer's self-attention and cross-attention mechanism, the dot product process between the query, key, and value matrices can be likened to a DA process. 
Thus, we hypothesize that the transformer architecture has the potential to be applied beyond MP and feature transferring for MOT task, which can be adapted for DA. 

Additionally, current tracking-related transformer algorithms apply to the case with a single sensor modality input (usually images), in a stationary position, with a high sampling frequency (usually 30 Hz), and for a single object category (the human class) \cite{dendorfer2020mot20}.
But for autonomous driving, tracking algorithms can operate with multi-modality sensor inputs (e.g., images and LiDAR-based point clouds), on a moving ego vehicle, with relatively low sampling frequency (10 Hz) and for multiple object categories (classes of pedestrian, car, truck, etc.). To the authors' best knowledge, no existing transformer tracking algorithms handle such intricate situations effectively.

Based on the above-mentioned assumptions and issues, we raise three questions:
(1) Can a transformer-based DA algorithm be applied for simultaneous MOD and MOT under an autonomous driving environment? If so, can a DA algorithm suffice without explicit MP and state estimation processes? 
(2) How to handle the multiple sensor inputs for DA? 
(3) Is it possible to enhance the detection performance through history-endowed tracking features? 
To answer these questions, we propose a novel end-to-end transformer-based algorithm (MotionTrack) for simultaneous MOD and MOT with LiDAR and image inputs (Figure \ref{fig:demo1}). 
MotionTrack algorithm utilizes a modified transformer to achieve DA and another transformer to update potential object features from the tracking information. 
The proposed algorithm is tested and evaluated through the nuScenes dataset, which achieves 2-3x higher AMOTA results than the other baseline algorithms, and is on par with popular tracking solutions, such as the probabilistic 3D Kalman filter. 
The contributions of this paper include: 
\begin{itemize}[noitemsep,nolistsep]
\item Designing a transformer-based module for DA with multi-modality sensor inputs to achieve tracking without MP and state estimation. 
\item Developing a query enhancement module (QEM) to improve detection performance by combining the history tracking features. 
\item Establishing a baseline for an end-to-end transformer MOD and MOT algorithm in the autonomous driving environment. 
\end{itemize}
To the best of our knowledge, this is the first end-to-end transformer algorithm for simultaneous MOD and MOT with multi-modality sensor inputs in an autonomous driving environment. 
We emphasize that the objective of this paper is to investigate the feasibility and establish a baseline rather than to achieve state-of-the-art (SOTA) results for tracking tasks with the nuScenes dataset; which would require further improvements on top of our baseline.

\section{Related Work}
Although objects move in a three-dimensional physical space, MOT can be performed to track objects in 3D or in 2D such as in an image \cite{RAKAI2022116300}. 
Inputs to MOT can be 2D such as images, or 3D such as Lidar point clouds, or both. 
MOT can employ traditional methods for MP, filtering, and DA, and it can employ neural networks to achieve desired goals for tracking.

\begin{figure*}[h]
\centering
\includegraphics[width=11cm]{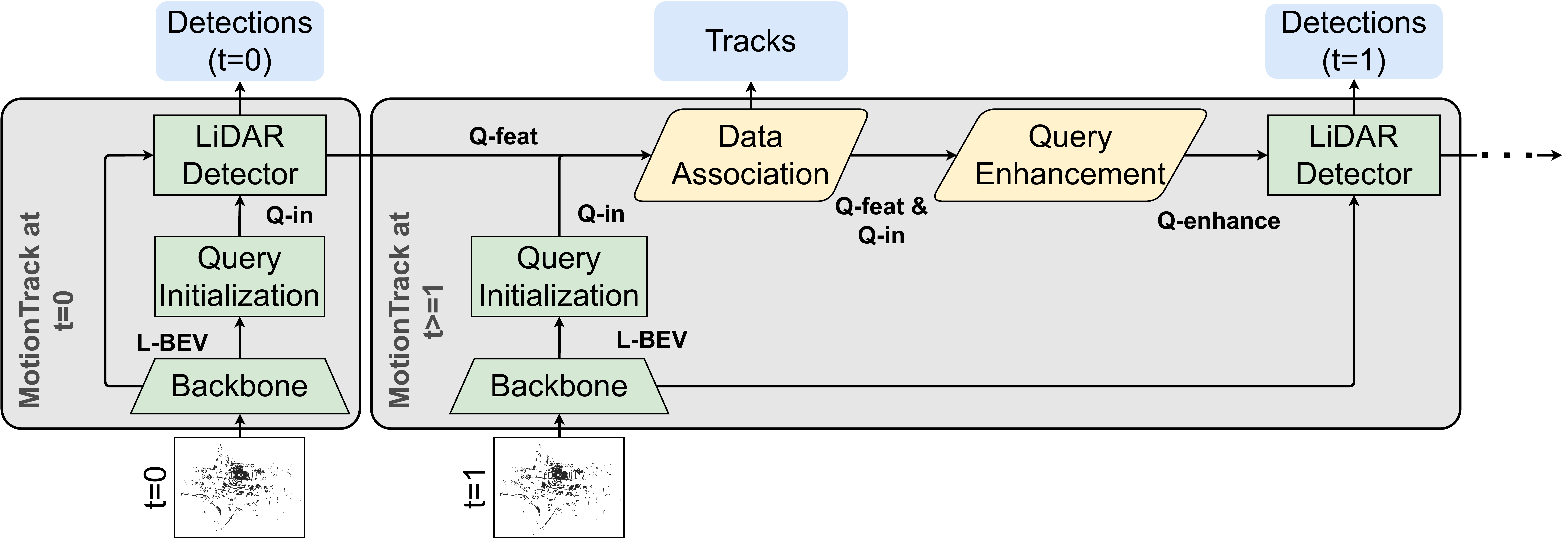}
\caption{MotionTrack Architecture with LiDAR Input. The green blocks represent the detection modules, the yellow blocks represent the tracking modules, and the blue blocks are the detection and tracking results.}
\label{fig:architecture1}
\end{figure*}

\begin{figure*}[h]
\centering
\includegraphics[width=11cm]{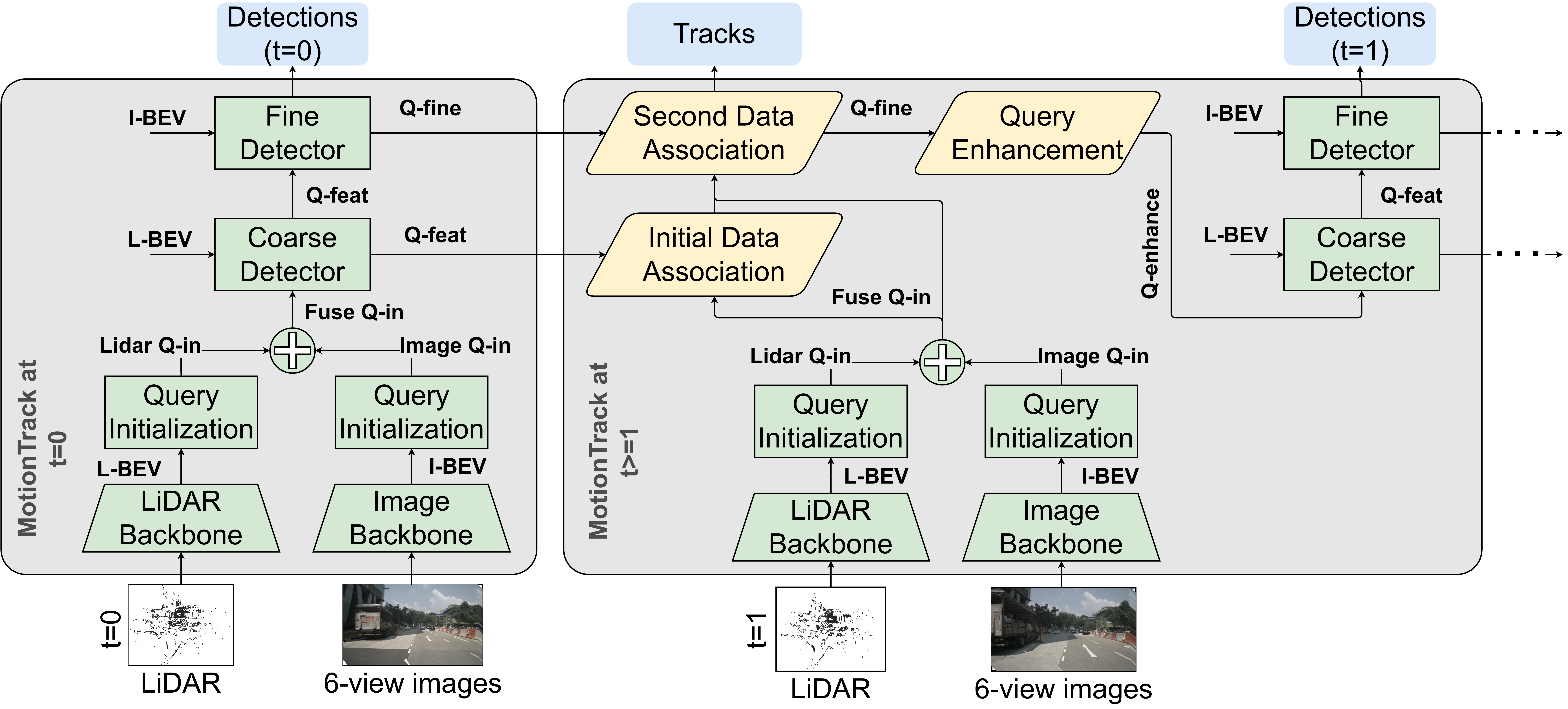}
\caption{MotionTrack Architecture with LiDAR-Image Input. The overall architecture is similar to the Figure \ref{fig:architecture1}, while there is an extra object association module for the fine detector results.}
\label{fig:architecture2}
\end{figure*}

\subsection{2D MOT}
2D MOT usually uses images and object states as input \cite{Zhang2021ByteTrackMT, Sharma2018BeyondPL, Schulter2017DeepNF, Wang2019TowardsRM, Tokmakov2021LearningTT, Bergmann2019TrackingWB}. 
Most 2D algorithms leverage the rich semantic information available in images and dense temporal features to accomplish MOT. 
However, image targets do not offer explicit position and motion information, strongly affects the 2D MOT performance. 
One of the most popular traditional 2D MOT methods is the Simply Online and Realtime Tracking (SORT) \cite{Bewley2016SimpleOA}, which employs a Kalman filter for MP and the Hungarian algorithm for DA across frames. 
The successor to SORT, namely DeepSORT \cite{Wojke2017SimpleOA}, modifies the DA algorithm to further improve tracking performance by utilizing a Mahalanobis distance assigner and a neural network-based appearance feature descriptor to assist DA between frames. 
Though traditional MOT algorithms are reliable and easy to deploy, it requires massive parameter tuning. 
Moreover, traditional MOT cannot deal with edge cases such as object occlusion. 

ML-based MOT algorithms are developed to solve the aforementioned issues in traditional tracking. 
CenterTrack \cite{zhou2020tracking} designs a convolutional neural network-based (CNN) MP module to estimate the heatmap displacements between 2 frames for DA. 
It achieves good performance on the MOT datasets while ID-switch and long-term tracking issues are yet to be solved. 

Besides CNN, transformer architecture becomes popular recently \cite{sun2020transtrack, 9879668, 10.1007/978-3-031-19812-0_38, Zhou2022GlobalTT, Zhu2021LookingBT, Xu2021TransCenterTW}, mainly thanks to its capability of global feature extraction and temporal feature aggregation. 
The global feature extraction processes all potential object queries simultaneously, while the temporal feature aggregation transfers past object features to the current frame as prior knowledge.  
Current popular transformer MOT algorithms are TrackFormer \cite{9879668}, MOTR \cite{10.1007/978-3-031-19812-0_38}, global tracking transformer (GTR) \cite{Zhou2022GlobalTT}, and MeMOT \cite{9880137}. 
The TrackFormer and MOTR follow the simultaneous detection and tracking paradigm by concatenating the detected objects' embedding with the proposed new-born query embedding. 
The GTR follows the tracking-by-detection paradigm, utilizing the self- and cross-attention mechanisms to associate objects among all input frames. 
The MeMOT comprises a DETR-based detector and three attention mechanisms to aggregate the object features from previous detections, before using an association solver for MOT. 
All these transformer-based algorithms demonstrate effective 2D MOT. 
However, these algorithms are designed for image-only tracking applications with a single class.

\subsection{3D MOT}
3D MOT algorithms, commonly applied for autonomous driving, take images, LiDAR sensor's point cloud, or LiDAR-image fusion data as inputs.

Image-based 3D MOT algorithms leverage dense appearance features for DA. 
AB3DMOT \cite{Weng2020AB3DMOTAB} employs a 3D Kalman filter and a Hungarian algorithm for tracking. 
\cite{Scheidegger2018MonoCamera3M} utilizes the Poisson multi-Bernoulli mixture tracking filter to achieve MOT with a single camera input. 
Besides traditional tracking methods, \cite{Pang2023StandingBP} and \cite{Zhang2022MUTR3DAM} introduce transformer-based 3D MOT models. 
Both use the transformer model in a manner of TrackFormer and MOTR, by concatenating tracked objects to the current frame for MOD and MOT.
\cite{Pang2023StandingBP} further aggregates temporal features to enhance tracking and detection. 
Despite dense appearance features, image-based 3D MOT algorithms' performance cannot compete with LiDAR-based methods due to a lack of explicit position and distance features. 

Most LiDAR-based algorithms focus on modeling the tracked objects' motion features to achieve MOT. 
SimTrack \cite{Luo2021ExploringS3}, and SimpleTrack \cite{Pang2021SimpleTrackUA} project the 3D features to BEV for feature extraction, where SimTrack utilizes neural networks to predict the object motion to achieve tracking, and SimpleTrack is focused on improving the association and motion model performance. 
\cite{Gwak2022MinkowskiTA}, inspired from the R-CNN detector \cite{7112511}, develops a track-align module to aggregates the track region-of-interests for MOT.

Currently, researchers are interested in LiDAR-image fusion as the image feature offers intensive appearance features, and the point cloud provides accurate distance and position features.
EagerMOT \cite{Kim2021EagerMOT3M} is a traditional 3D tracker that employs a two-stage association algorithm where the first stage is a standard Hungarian data association for 3D bounding boxes, and the second stage is an identical Hungarian assigner with 2D bounding boxes. 
JMODT \cite{Huang2021JointMD} is a joint detection and tracking algorithm that uses point clouds and images as inputs with a novel neural network-based object correlation and object association. 
CAMO-MOT \cite{Wang2022CAMOMOTCA} combines motion and appearances features to prevent false detection. 
In the meantime, they design a tracking cost matrix to prevent tracking occlusion, which achieves  the current SOTA algorithm at nuScenes dataset.
Even though numerous literatures, to the best of the authors' knowledge, there are no multi-modality, end-to-end transformer algorithms that exist for 3D MOT autonomous driving.

\section{PROPOSED METHODOLOGY}
MotionTrack is a simultaneous detection and tracking algorithm. 
For MOD, we employ the TransFusion model. 
For MOT, we design a transformer-based DA module. 
Furthermore, we develop a QEM to inherit the history temporal information for better detection performance.
The MotionTrack algorithm comprises two setups with different sensor configurations: LiDAR-only input (Figure \ref{fig:architecture1}) and LiDAR-image fusion inputs (Figure \ref{fig:architecture2}). 
Both setups contain an object detection module, a DA module, and a QEM. 
Due to the intricacy of the MOT problem, we define several terms for the MotionTrack (Table \ref{table1})

\begin{table}[]
\centering
\caption{MotionTrack terminology and acronym names.}
\scalebox{0.7}{
    \begin{tabular}{M|Q|S}
    \toprule
    \textbf{Terminology} & \multicolumn{1}{K|}{\textbf{Explanation}} & \textbf{Acronym}      \\\midrule
    {Coarse (LiDAR) Detector} & {A transformer decoder layer to extract features from the input queries for object detection. It is named as LiDAR detector in the LiDAR-only setup and coarse detector in the LiDAR-image setup.} & C-Det                 \\\midrule
    {Fine Detector} & {A transformer decoder layer to extract features from the input queries for object detection with LiDAR-image setup}                                                                     & F-Det                 \\\midrule
    {Query Input}   & {Queries generated based on heatmap results and use as the input to the coarse or LiDAR detector at the initial frame or the input to the QEM at the following frames}        & Q-in   \\\midrule
    {Query Features}            & {The output from the coarse or LiDAR detector, also used as the input to the fine detector and the DA module}     & Q-feat \\\midrule
    {Query Features Fine}       & {The output from the fine detector, also use as the input to the extra DA module}    & Q-fine  \\\midrule
    {Enhanced Query}            & {The enhanced queries generated from the QEM, used as the input to coarse or LiDAR detector} & Q-enhance   \\\midrule
    {LiDAR BEV}                   & {The LiDAR BEV features}    & L-BEV \\\midrule
    {Image BEV}                   & {The image BEV features}    & I-BEV   \\\midrule
    {query cross-attention}   & {A cross attention layer to update the query features from previous frame}  & Q-cross  \\\midrule
    {head cross-attention}    & {A cross attention layer to update the heading angle features to query features}  & H-cross  \\\bottomrule
    \end{tabular}
}
\label{table1}
\end{table}

\subsection{MotionTrack Detector Module}
MotionTrack's detector is TransFusion \cite{Bai2022TransFusionRL}, which flexibly supports LiDAR-only and LiDAR-image setups.

The LiDAR-only setup supports two implementations: PointPillar \cite{Lang2018PointPillarsFE} as a cost-efficient model; or VoxelNet \cite{Zhou2017VoxelNetEL} for high performance.
The extracted features are transformed into BEV features (L-BEV) for object query initialization. 
Then, we generate heatmaps based on the L-BEV to determine the initial location of the Q-in. 
After the heatmap generation, an one-layer transformer decoder takes the L-BEV and Q-in as the inputs to extract Q-feat. 
Finally, the Q-feat is fed into prediction head layers for object detection.

Compared to the LiDAR-only setup, the LiDAR-image one shares the same LiDAR backbone, outputting L-BEV and LiDAR's Q-in. 
In the image branch, ResNet-50 \cite{He2015DeepRL} backbone first extracts features from 6-view RGB images, and projects them to BEV (I-BEV). 
Then, we generate the image heatmap, before fusing the image's and LiDAR's Q-in together.
The detection in the LiDAR-image setup has two transformer decoder layers: one is C-Det, which consumes L-BEV and Q-in; the other one is F-Det, which uses I-BEV and the C-Det's output (Q-feat) as the inputs.
Finally, the F-Det's output is fed into the prediction head layers.
Details about the detector module can be found in \cite{Bai2022TransFusionRL}. 

\subsection{Transformer-based Association Module}

Common MOT algorithms perform motion prediction (MP) and data association (DA) in sequence.
The core of MOT is DA, as it enables the connection of objects between frames, while the MP serves to support the DA process. 
The outcome of a DA algorithm is whether current frame objects are tracked or new-born objects, and whether the previous frame's objects are disappeared, namely dead objects \cite{Granstrm2016ExtendedOT}.
Here we investigate a DA design through the transformer architecture without explicit MP.

\textbf{\textit{Transformer DA Inspiration:}} The self- and the cross-attention mechanism are the core of transformers.
For both mechanisms, the attention function is $softmax(Q*K^T)*V$ where $Q$, $K$, and $V$ are known as the query, key, and value matrices learned from the inputs. 
The $Q$, $K$, and $V$ matrices are learned from the same inputs for the self-attention (input-A), while the $Q$ are learned from a different input with the $K$ and $V$ matrices for the cross-attention \cite{vaswani2017attention}.   
The essence of the attention function is to update the value matrix based on a cost matrix ($softmax(Q*K^T)$) obtained from query and value matrices.

The cost matrix inspires us that such an attention mechanism can be used to calculate the affinity between the tracks and the observations.
However, the experiment results indicate that directly applying the attention mechanism is not ideal because the softmax activation function causes the gradient vanishing issue during training. 
Therefore, instead of the softmax activation function, we directly compute the cost matrix with a dot product between the observation and tracks features, which shows excellent association performance.
Furthermore, we find that the attention mechanism performs well on updating object features from the previous frame to the current frame because the $softmax(\frac{Q*K^T}{C})$ can easily learn to filter redundant features and preserve necessary features from the previous frame.
Based on these findings, MotionTrack's association module employs transformer architecture to update the previous frame's objects features and uses a dot product computation for DA.

The MotionTrack comprises two DA module configurations for LiDAR-only and LiDAR-image inputs.
Each DA module outputs independent tracklets estimations. 
The differences are that the LiDAR-image inputs comprise an extra DA module.

\begin{figure}[t!]
\centering
\includegraphics[width=8.3cm]{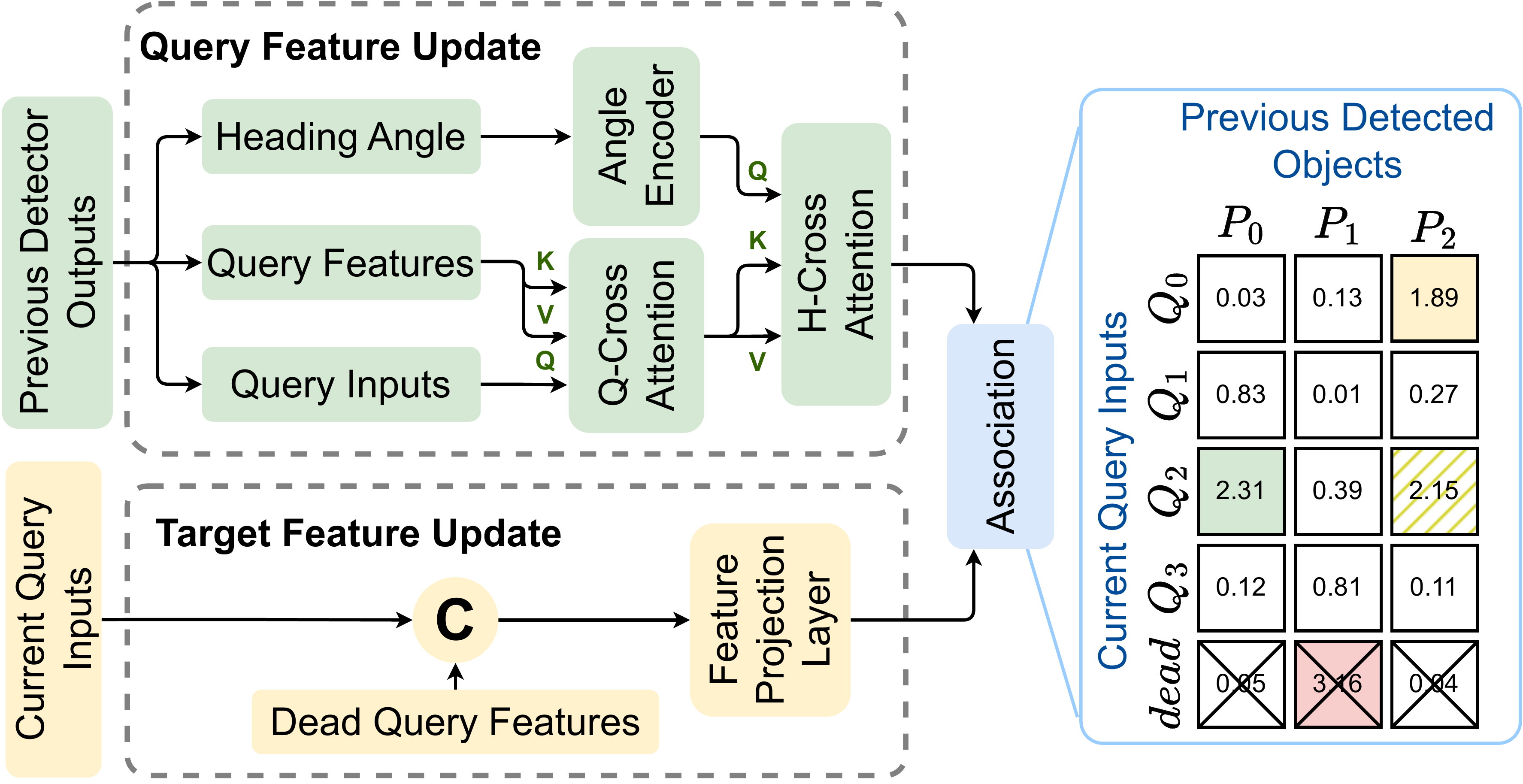}
\caption{Object Association Module. The cross marked rectangle represents dead objects. When a duplicate association occurs, the association outcomes are determined by greedy matching (e.g. $Q_2$ versus $P_0$ and $P_2$).}
\vspace{-0.1cm}
\label{fig:asso}
\end{figure}

\textbf{\textit{DA Module:}} 
The DA module contains three steps: query feature update, target feature update, and query-target feature association (Figure \ref{fig:asso}). 

The query feature update process is aimed at establishing and enhancing detected object features from previous frames.
It takes the previous frame's detected Q-feat, Q-in, and the objects' heading angles as the input, passing through two cross-attention layers (H-cross \& Q-cross) to update the detected objects' features.
The objective of the Q-cross is to update the appearance feature for detected objects. Appearance features are important for DA since the association is determined by the similarity between the appearance features.
The H-cross is designed to inherit the objects' motion features. 
Since most single-frame detection algorithms cannot estimate object movement accurately, such as velocity acceleration, etc., heading angle is one of the most important motion features to introduce for DA. 

As for the target update module, it aims to refine the current frame's object candidate features for DA.
The target update module simply takes the current frame's Q-in as the input and passes through a two-layer multilayer-perceptron (MLP) to prepare for DA because feature updating is not necessary for current frame's features. 
Since the previously detected objects might disappear at the current frame, an empty vector (filled with zeros) is concatenated to the Q-in, which we call "dead query features," to represent the disappeared object queries. 

Finally, the previous frame's updated query features are associated with the current frame's refined target features through a dot product process.
The output from the dot product operation is an $N$ by $M+1$ matrix where $N$ is the number of detected objects from the previous frame, and $M$ is the number of object queries in the current frame. 
A higher value in the matrix (association score) indicates a higher possibility of an association. 
During the training phase, we treat the association process as a classification task and compute the loss between the object association module estimated results and the ground truth results with the cross-entropy loss function. 
During the evaluation phase, we apply the same method as the training phase but employ a greedy-based search method after the object association module to prevent duplicate association.

\textbf{\textit{Extra DA Module:}}
The extra DA module is designed for the LiDAR-image fusion detector. 
The architecture design is the same as the LiDAR-only DA module.
The difference is that the query feature update model uses the previous frame's Q-fine, Q-feat, and detected objects' heading angles as the input.
Moreover, the extra DA requires a further decision-making step during the evaluation phase.
When both DA modules estimate the same tracklet results, the final association is determined by the extra association module. When there is a conflict between the two association modules, the final association results are determined by the highest association score.

\subsection{Query Enhancement Module}
\begin{figure}[h]
\centering
\includegraphics[width=8.2cm]{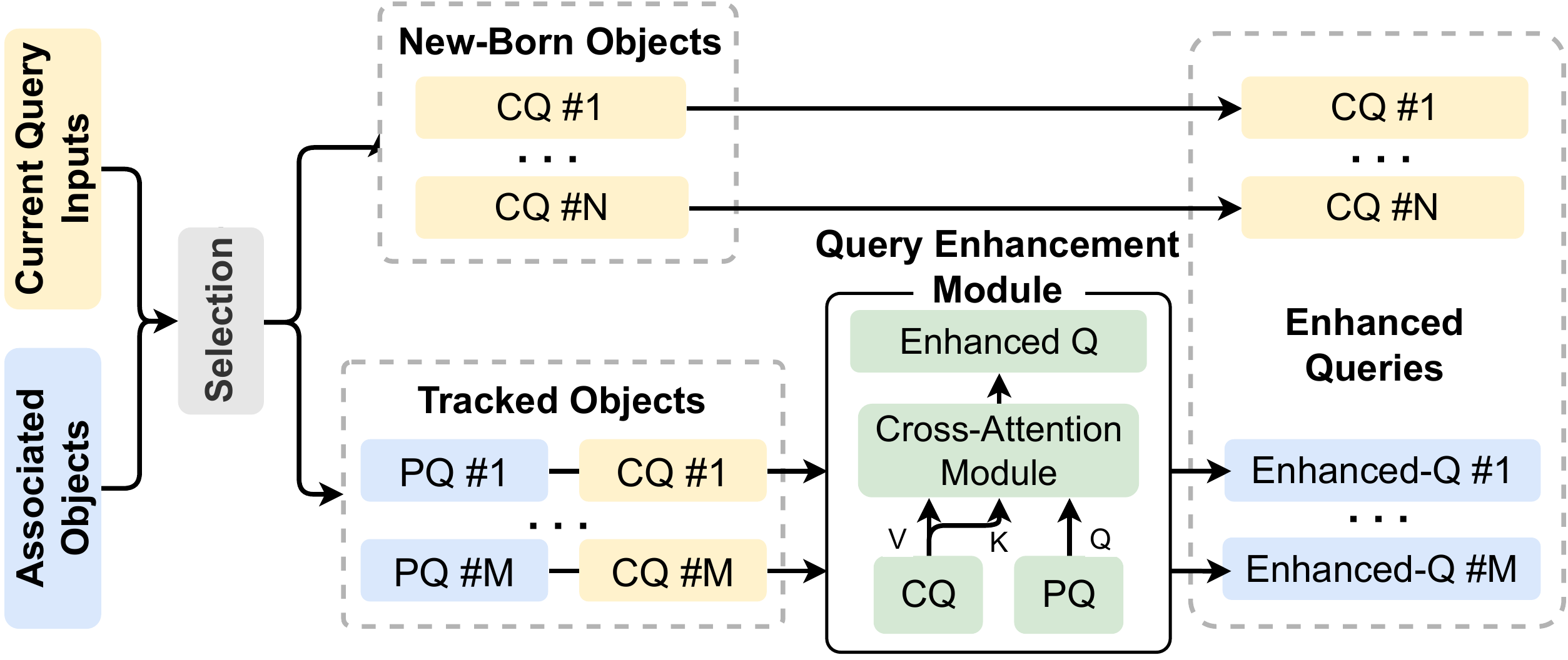}
\caption{Query Enhancement Module. CQ represents the current frame's queries, PQ represents the previous frame's queries, and Enhanced-Q represents the enhanced queries. For the new-born objects, the query inputs are directly used as the query input to the transformer decoder.}
\vspace{-0.3cm}
\label{fig:query_enhance}
\end{figure}

The objective of the QEM is to imbue the current frame's Q-in with the previous frame's Q-feat or Q-f2 to improve the detection performance, and the overall architecture is shown in Figure \ref{fig:query_enhance}. 
In the QEM, the previous frame's Q-feat or Q-fine, and the current frame's Q-in are passed into a cross-attention layer to aggregate the previous frame's features to the current frames' corresponding queries. In this way, the history temporal information is aggregated to the current frame. 
Furthermore, such a cross-attention mechanism can prevent unnecessary or even misleading information contaminate the current frame's features through model learning. J. Koh, et al. have conducted a similar operation for temporal information aggregation. \cite{Koh2021Joint3O}

\subsection{Training And Evaluation Model Differences}
The differences between the training phase and evaluation phase are the number of input frames, object detection decision-making, and object association processing.
For the training phase, the number of input frames is set to 2, while for the evaluation phase, the number of input frames is 1 at a time. 
As for the object detection decision-making, we follow the same procedure with \cite{Bai2022TransFusionRL}. For the object association process, the association results are processed in a manner of classification task. During training, the association module's outputs are compared with the ground truth associations results and the loss are computed with a cross-entropy equation. In the LiDAR-image setup, both association modules' output tracklets are compared with the ground truth associations results for loss computation. In the evaluation phase, the association results are further processed with a greedy-based search algorithm to prevent duplicate associations.
The overall training procedures can be summarized as
\begin{enumerate}[noitemsep, nolistsep,itemindent=3pt] 

\item Train the TransFusion model by following the instructions from \cite{Bai2022TransFusionRL}. 
\item Train the MotionTrack model with a DA module based on the TransFusion model's checkpoint. 
In this step, all parameters are frozen except the DA module. 
\item Train the MotionTrack model with both the DA and the QEM. 
In this process, all parameters are learned except the backbones.
\end{enumerate}

\section{RESULTS \& DISCUSSIONS}

The MotionTrack model and its variants are evaluated by the nuScenes dataset. 
The results comprise the detection and the tracking results. 

\subsection{Imeplementation Details}
The input data of the MotionTrack is a 360-degree point cloud and 6 RGB images that capture the ego vehicle's surrounding views. 
As for the training and validation set split, we use the official split of the training, validation, and evaluation dataset. 
All training experiments are conducted with eight A100 80GB GPUs. As for the training parameters, we use the AdamW optimizer with one cycle learning rate policy, with a max learning rate 0.001, weight decay 0.01, and momentum 0.85 to 0.95. 
Since the number of tracked objects varies across all training samples, we set the training batch size to be one per GPU.
\subsection{MotionTrack Detection Results}
The MotionTrack detection results of the nuScenes dataset are presented in Table \ref{table2}. 
As for the detection results, the VoxelNet backbone performs better than the PointPillar since the former voxelized the original point cloud data with smaller segments.
Furthermore, the VoxelNet backbone achieves better results on small objects such as pedestrians, motorcycles, and bicycles. 
The MotionTrack-Voxel's AP performances are 13.03\%, 44.87\%, and 112.45\% higher than the MotionTrack-Pillar's on pedestrian, motorcycle, and bicycle. 
Even though the MotionTrack-Voxel outperforms the MotionTrack-Pillar in all object categories, both models' performance for the car class is similar because the objects size is relatively large and the number of training samples is large. 
Moreover, the MotionTrack-Pillar is more efficient where the model size is 36.4\% smaller than the MotionTrack-Voxel. 

\begin{table}[]
\caption{Detection results on nuScenes validation dataset.}
\centering\scalebox{0.7}{
\begin{tabular}{c|c|ccc}
\toprule
 \textbf{Model}       & {\textbf{Classes}} &{\textbf{AP} $\uparrow$} & {\textbf{ATE} $\downarrow$} & {\textbf{AAE} $\downarrow$} \\ \midrule
\multirow{7}{*}{MotionTrack-Pillar}    & Bicycle                     & 0.27                  & 0.25                   & 0.02                   \\
                                        & Bus                         & 0.64                  & 0.40                   & 0.31                    \\
                                        & Car                         & 0.83                  & 0.21                   & 0.19                    \\
                                        & Motorcycle                  & 0.50                  & 0.25                   & 0.16                   \\
                                        & Pedestrian                  & 0.78                  & 0.15                   & 0.09                   \\
                                        & Trailer                     & 0.36                  & 0.57                   & 0.21                   \\
                                        & Truck                       & 0.50                  & 0.39                   & 0.22                   \\ \midrule
\multirow{ 7}{*}{MotionTrack-Voxel} & Bicycle                     & \textbf{0.58}                 & 0.16                   & 0.01                   \\
                                    & Bus                         & \textbf{0.73}                  & 0.35                   & 0.26                   \\
                                    & Car                         & \textbf{0.87}                 & 0.17                   & 0.20                   \\
                                    & Motorcycle                  & \textbf{0.72}                 & 0.20                   & 0.23                   \\
                                    & Pedestrian                  & \textbf{0.88}                & 0.13                   & 0.08                   \\
                                    & Trailer                     & \textbf{0.44}                  & 0.52                   & 0.17                   \\
                                    & Truck                       & \textbf{0.60}                  & 0.33                   & 0.24                   \\ \bottomrule
\end{tabular}}
\label{table2}
\end{table}

\subsection{MotionTrack Tracking Results}
We compared the MotionTrack with other baseline tracking algorithms, such as the AB3DMOT, CenterTrack, and Probabilistic 3D Kalman filter in the nuScenes test dataset (Table \ref{table3}). 
The selected algorithms contain both ML-based and traditional tracking methods. 
The main evaluation metric for MOT is AMOTA, which is integrals over the MOTA metric using n-point interpolation. The AMOTA equation is available in \cite{Caesar2019nuScenesAM} 


According to the AMOTA results, the proposed MotionTrack baseline is 3.7x higher than the CenterTrack, 2.3x higher than the AB3DMOT, and on par with the probabilistic 3D Kalman filter model. 
This result proves that a simple transformer-based association algorithm can achieve MOT under an autonomous driving environment. 
Furthermore, the improvement compared with the traditional Kalman filter-based method and the tracking-by-detection paradigm indicates that the simultaneous tracking and detection paradigm with transformer architecture has huge potential. 
According to the comparison results, we found that the high object ID switching is the reason that affects the overall tracking performance. 
The reason cause such an issue is that we don't introduce a carefully designed tracking management algorithm during the inference time. 

Table \ref{table4} tabulates the MotionTrack's tracking results among all categories. 
According to Table \ref{table4}, the AMOTA results are proportional to the number of object samples. 
The reason is that the transformer DA module requires numerous samples to learn the objects' features. 
Furthermore, Table \ref{table4}'s results indicate that larger objects (car, bus, and truck) exhibit better performance than small objects, especially for the MotionTrack with LiDAR-only input. 
Even though the multi-sweep LiDAR input method is applied, the point cloud is still sparse for small objects. 
Therefore, the association accuracy is decreased due to poor object features. 
Such poor tracking performance is solved by the MotionTrack with image-LiDAR input because of the dense appearance features obtained from images. 
According to Table \ref{table4}, the pedestrian and motorcycle AMOTA results for MotionTrack with image-LiDAR input are 5.6\% and 11.2\% higher than the MotionTrack with LiDAR-only.       

\begin{table*}[]
\caption{MotionTrack compared with other baselines on nuScenes test dataset. Bold and underlined text represent the top and second results.}
\centering
\small\scalebox{0.7}{
\begin{tabular}{l|cccccccccc}
\toprule
\textbf{Model Name}          & \textbf{Modality} & \textbf{AMOTA $\uparrow$} & \textbf{AMOTP $\downarrow$} & \textbf{MOTA $\uparrow$} & \textbf{MOTAR $\uparrow$} & \textbf{MOTAP $\downarrow$}  & \textbf{FAF $\downarrow$}  & \textbf{MT $\uparrow$}  & \textbf{ML $\downarrow$}  & \textbf{IDS $\downarrow$}  \\ \midrule
PointPillar+AB3DMOT     & L                                         & 0.03 & 1.70 & 0.05 & 0.24 & 0.82 & 220.9 & 480 & 5332 & \underline{7548} \\
Prob-3DKalman           & L                                         & \underline{0.55} & \textbf{0.80} & 0.46 & \underline{0.77} & 0.35 & 54.5 & \textbf{4294} & 2184 & \textbf{950} \\
CenterTrack             & L+C                                       & 0.11 & 0.99 & 0.09  & 0.27 & 0.35 & 206.6  & 1308 & 3739 & 7608 \\
AB3DMOT                 & L+C                                       & 0.15 & 1.50 & 0.28  & 0.55 & 0.15 & 55.8   & 1006 & 4428 & 9027 \\ \midrule[0.3pt]
\textbf{MotionTrack}$_{\textrm{{Pillar-L}}}$   & L                  & 0.42 & 1.01 & 0.385  & 0.74 & 0.34 & 42.8  & 3850 & 2758 & 10139 \\
\textbf{MotionTrack}$_{\textrm{{Voxel-L}}}$    & L                  & 0.51 & 0.99 & \underline{0.48}  & \textbf{0.83 }& \underline{0.30} & \underline{28.4}  & 3723 & 1567 & 9705 \\
\textbf{MotionTrack}$_{\textrm{{Pillar-LC}}}$  & L+C                & 0.45 & 0.90 & 0.48  & 0.59 & 0.31 & 32.7  & 3014 & 1815 & 9943 \\
\textbf{MotionTrack}$_{\textrm{{Voxel-LC}}}$   & L+C                & \textbf{0.55} & \underline{0.871} & \textbf{0.49}  & 0.77 & \textbf{0.26} & \textbf{22.4}  & \underline{4211} & \textbf{1321} & 8716 \\ \bottomrule
\end{tabular}}
\label{table3}
\end{table*}

\begin{table*}[]
\caption{MotionTrack tracking results among all categories on nuScenes validation dataset.}
\centering\small\scalebox{0.7}{
\begin{tabular}{ccccccccccccc}
\toprule
\multicolumn{12}{c}{\textbf{MotionTrack-Voxel-LiDAR}}                                                                                              \\ \midrule
\multicolumn{1}{c|}{\textbf{Classes}}    & \textbf{AMOTA $\uparrow$} & \textbf{AMOTP $\downarrow$} & \textbf{Recall $\uparrow$} & \textbf{MOTAR $\uparrow$} & \textbf{GT}    & \textbf{MOTA $\uparrow$}  & \textbf{MOTP $\downarrow$}  & \textbf{MT $\uparrow$}   & \textbf{ML $\downarrow$}  & \textbf{FAF $\downarrow$}  & \textbf{IDS $\downarrow$}  \\ \midrule
\multicolumn{1}{l|}{Bicycle}    & 0.40 & 0.22 & 0.48  & 0.90 & 1993  & 0.43  & 0.21 & 28   & 68  & 6.5  & 11        \\
\multicolumn{1}{l|}{Bus}        & 0.73 & 0.61 & 0.79  & 0.90 & 2112  & 0.68 & 0.40   & 61   & 10  & 9.6  & 74        \\
\multicolumn{1}{l|}{Car}        & 0.85 & 0.19 & 0.74  & 0.93 & 58317 & 0.61 & 0.21  & 1856 & 637 & 45.2 & 4631    \\
\multicolumn{1}{l|}{Motorcycle} & 0.60 & 0.47 & 0.67  & 0.92 & 1977  & 0.56 & 0.27 & 58   & 21  & 7    & 126      \\
\multicolumn{1}{l|}{Pedestrian} & 0.80 & 0.22 & 0.75  & 0.87 & 25423 & 0.60 & 0.23 & 923  & 319 & 50.9 & 15678    \\
\multicolumn{1}{l|}{Trailer}    & 0.37 & 0.57 & 0.56   & 0.72 & 2425  & 0.36 & 0.57 & 53   & 50  & 33.7 & 143    \\
\multicolumn{1}{l|}{Truck}      & \textbf{0.57} & 0.34 & 0.63  & 0.75 & 9650  & 0.46 & 0.34 & 210  & 150 & 39.1 & 204   \\ \midrule
\multicolumn{12}{c}{\textbf{MotionTrack-Voxel-ImageLiDAR}}                                                                                         \\ \midrule
\multicolumn{1}{c|}{\textbf{Classes}}    & \textbf{AMOTA $\uparrow$} & \textbf{AMOTP $\downarrow$} & \textbf{Recall $\uparrow$} & \textbf{MOTAR $\uparrow$} & \textbf{GT}    & \textbf{MOTA $\uparrow$}  & \textbf{MOTP $\downarrow$}  & \textbf{MT $\uparrow$}   & \textbf{ML $\downarrow$}  & \textbf{FAF $\downarrow$}  & \textbf{IDS $\downarrow$}  \\ \midrule
\multicolumn{1}{l|}{Bicycle}    & \textbf{0.41} & 0.20 & 0.50  & 0.95 & 1993  & 0.51 & 0.21 & 28   & 65  & 6.3  & 11     \\
\multicolumn{1}{l|}{Bus}        & \textbf{0.78} & 0.55 & 0.85  & 1.06 & 2112  & 0.76 & 0.44 & 71   & 10  & 8.0  & 62     \\
\multicolumn{1}{l|}{Car}        & \textbf{0.86} & 0.15 & 0.83  & 0.99 & 58317 & 0.66 & 0.19 & 2150 & 589 & 45.1 & 4403   \\
\multicolumn{1}{l|}{Motorcycle} & \textbf{0.66} & 0.45 & 0.72  & 1.07 & 1977  & 0.64 & 0.17 & 90   & 19  & 6.6  & 111    \\
\multicolumn{1}{l|}{Pedestrian} & \textbf{0.84} & 0.24 & 0.80  & 1.04 & 25423 & 0.65 & 0.22 & 1073 & 304 & 45.5 & 15081  \\
\multicolumn{1}{l|}{Trailer}    & \textbf{0.38} & 0.54 & 0.59  & 0.86 & 2425  & 0.42 & 0.52 & 54   & 49  & 32.4 & 139    \\
\multicolumn{1}{l|}{Truck}      & 0.56 & 0.31 & 0.71  & 0.87 & 9650  & 0.54 & 0.28 & 219  & 148 & 31.9 & 193    \\ \bottomrule
\end{tabular}}
\label{table4}
\end{table*}

\subsection{Ablation Studies}
We conduct two ablation studies to validate the importance of the transformer-based DA module and the QEM.
\begin{table}[]
\centering
\small
\caption{DA module ablation study on nuScenes validation dataset.}
\scalebox{0.8}{
\begin{tabular}{M|MSS}
\toprule
\textbf{Model}                                    & \textbf{Module }         & \textbf{AMOTA}  & \textbf{MOTA}   \\ \midrule
\multirow{2}{*}{\textbf{MotionTrack}$_{\textrm{{Voxel-L}}}$} & w/ Transformer  & 0.62  & 0.53  \\
                                         & w/o Transformer  & 0.22  & 0.20  \\ \midrule
\multirow{2}{*}{\textbf{MotionTrack}$_{\textrm{{Voxel-LC}}}$} & w/ Transformer  & 0.69  & 0.61  \\ 
                                         & w/o Transformer & 0.23  & 0.20 \\ \bottomrule
\end{tabular}}
\label{table5}
\end{table}

\begin{table}[]
\centering\small
\caption{QEM ablation study for car class on nuScenes validation dataset.}\scalebox{0.8}{
\begin{tabular}{M|MSS}
\toprule
Model                                       & Module            & mAP & AMOTA \\ \midrule
\multirow{2}{*}{\textbf{MotionTrack}$_{\textrm{{Voxel-L}}}$}    & w/ Query Enhance  & 0.87     & 0.85       \\
                                            & w/o Query Enhance & 0.81     & 0.85       \\ \midrule
\multirow{2}{*}{\textbf{MotionTrack}$_{\textrm{{Voxel-LC}}}$} & w/ Query Enhance  & 0.88     & 0.93       \\
                                            & w/o Query Enhance & 0.86     & 0.93       \\ \bottomrule
\end{tabular}}
\label{table6}
\vspace{-0.1cm}
\end{table}

\textbf{\textit{Transformer-based Data Association}}
The objective of the transformer module is to refine and update the object features based on previous frame objects' appearance, position, and heading angle features. 
Furthermore, a simple dot product computation cannot accurately associate objects between consecutive frames.
In the DA module ablation study, we conduct two experiments: 
(1) A MotionTrack model contains the transformer process so that both the previous frame's object features and the current frame's queries are processed through the transformer module before dot product association 
(2) A MotionTrack model without transformer-based DA so that the previous frame's objects' features and the current frame's queries are directly associated through dot product. 
According to Table \ref{table5}, the transformer-based DA process is necessary since the AMOTA is almost 3x higher than the one without the transformer process.
Due to the hardware limitations and the complex ego vehicle's dynamic behavior, autonomous vehicle's data quality is inferior to other tracking-related data such as the MOT16. 
For instance, the sampling frequency of the nuScenes dataset is only 2 Hz, and the target objects' local movement is dependent on the ego vehicle's motion.

\textbf{\textit{QEM}}
The QEM helps with the detection performance while the tracking performance does not significantly improve. 
The objective of the QEM is to employ temporal features for object detection. 
According to Table \ref{table6}, the proposed QEM improves the detection performance by 6\% and 3\% for the MotionTrack with LiDAR-only input and the image-LiDAR input. 
Even though better detection results, the tracking performance is not improved accordingly, which against conventional intuition.
According to our analysis, when detection performance improved, the ID switch is increased correspondingly.
Since the MotionTrack only associates objects between consecutive frames without further track management (e.g. disappear objects buffer), increasing the number of detected bounding boxes increases the possibility of wrong objects associations across frames.  
Therefore, even though the proposed query enhancement is helpful with object detection but the tracking performance does not improve accordingly. 
We believe that QEM can help with the overall tracking performance with a well-designed track management algorithm and longer input frames during the training phase.

\subsection{Discussions and Potential Improvements}
Although MotionTrack is more performant than other baselines, we are aware that it cannot compete with SOTA algorithms such as the ImmortalTrack \cite{Wang2021ImmortalTT}, CAMO-MOT \cite{Wang2022CAMOMOTCA}, and ByteTrack \cite{Zhang2021ByteTrackMT}. 
Nevertheless, this paper's objective is to provide a good starting point for multi-modality end-to-end transformer-based MOT research. 
Below, we discuss MotionTrack's four potential improvements.

First, longer input frames for training can improve the robustness of the association against occlusion.
MotionTrack set the input frame to 2, which do not contain relatively long temporal information to simulate the actual tracking process. 
This issue is also reflected by the poor object ID switch results. With a longer number of input frames, the association algorithm can learn associations from more complex cases, such as occlusion. 

Second, a better track management module integrated with model inference can help perform object reidentification (ReID).
Currently, MotionTrack only considers DA between consecutive frames. 
If a previous frame's object is failed to associate with any current frame's object, that previous frame's object is directly considered a dead object. 
Existing algorithms provide a "disappear object" buffer during the inference phase so that the non-associated objects can be associated again with future frames to prevent tracking loss due to occlusion. MotionTrack's MOT performance can be improved after introducing such similar design.

New tracking-oriented data augmentation methods is the third improvement we propose for potential improvement.
MotionTrack's current data augmentation methods are mainly designed for MOD, such as random flips, rotations, and scales. 
We realize that there are several augmentation techniques, such as randomly dropping tracked objects and adding false positive objects, but these techniques only marginally improve the MotionTrack's tracking performance. 
To further improve the robustness of the DA and QEM, more effective data augmentation algorithms are required.

The last potential improvement is to properly process the MotionTrack with a larger batch size used in model training, in order to speed things up without dramatically increasing the memory footprint. 
This issue also occurred with other end-to-end tracking algorithms, such as the MOTR and the MeMOT algorithm. 
The reason is that the number of detected objects are varied among different frames, which causes the dimension of the previous frame's object features to be inconsistent.
Currently, a common solution is to concatenate zero vectors to represent empty objects so that the object features' dimensions are the same across training samples. 
However, this method causes the memory footprint and model sizes increase, especially for transformer architecture, which can potentially exceed certain GPUs memory cap.
Therefore, a design that can process the MotionTrack with higher batch sizes without wasting the memory footprint can speed up the training.

In summary, the proposed MotionTrack is a baseline for multi-modality end-to-end transformer-based MOT. 
Our results indicate a huge potential for transformer-based MOT. 
\section{CONCLUSIONS}
 
This paper proposes a novel simultaneous detection and tracking baseline algorithm, MotionTrack, with multi-modality sensors inputs under autonomous driving environment. 
MotionTrack proves that the transformer-based algorithm is suitable for MOT under the autonomous driving environment. 
Furthermore, MotionTrack validates that the self- and the cross-attention mechanism is capable of objects’ association with multiple classes. 
Finally, we propose a transformer-based query update algorithm, QEM, to refine the potential object queries from history frames to improve the overall detection performance. 

MotionTrack’s tracking results outperform other baseline algorithms on the nuScenes dataset. 
The current drawbacks and potential improvements to MotionTrack are elaborated in the results section. 
We believe the MotionTrack can be used as a new baseline algorithm for future deep learning-based end-to-end tracking algorithms in the autonomous driving environment.

{\small
\bibliographystyle{ieee_fullname}
\bibliography{literature_review}
}

\end{document}